# Inter-database validation of a deep learning approach for automatic sleep scoring

Diego Alvarez-Estevez[1] and Roselyne M. Rijsman[1]

(1) Sleep Center, Haaglanden Medisch Centrum, Lijnbaan 32, 2512VA, The Hague, The Netherlands

**Abstract**
In this work we describe a new deep learning approach for automatic sleep staging, and carry out its validation by addressing its generalization capabilities on a wide range of sleep staging databases. Prediction capabilities are evaluated in the context of independent local and external generalization scenarios. Effectively, by comparing both procedures it is possible to better extrapolate the expected performance of the method on the general reference task of sleep staging, regardless of data from a specific database. In addition, we examine the suitability of a novel approach based on the use of an ensemble of individual local models and evaluate its impact on the resulting inter-database generalization performance. Validation results show good general performance, as compared to the expected levels of human expert agreement, as well as state-of-the-art automatic sleep staging approaches.

**1. INTRODUCTION**

Sleep staging is one of the most important tasks during the clinical examination of polysomnographic sleep recordings (PSGs). A PSG records the relevant biomedical signals of a patient in the context of Sleep Medicine studies, representing the basic tool for the diagnosis of many sleep disorders. Sleep staging characterizes the patient's sleep macrostructure leading to the so-called hypnogram. The hypnogram plays also a fundamental role for the interpretation of several other biosignal activities of interest, such as the evaluation of the respiratory function, or the identification of different body and limb movement [1] [2]. Current standard guidelines for sleep scoring carry out segmentation of the subject's neurophysiological activity following a discrete 30s-epoch time basis. Each epoch can be classified into five possible states (wakefulness, stages N1, N2, N3, and R) according to the observed signal pattern activity in the reference PSG interval. Specifically, for sleep staging, neurophysiological activity of interest involves monitoring of different traces of electroencephalographic (EEG), electromyographic (EMG) and electrooculographic (EOG) activity [1].

A typical PSG examination comprises 8 up to 24 hours of continuous signal recording, and its analysis is usually carried out manually by an expert clinician. The scoring process is consequently expensive and highly demanding, due to the involved clinician's time, and the complexity of the analysis itself. Moreover, the demand for PSG investigations is growing in relation with the general public awareness, motivated by clinical findings over the last years uncovering the negative impact that sleep disorders exert over health. This represents a challenge for the already congested sleep centers, with steadily increasing waiting lists.

Automatic analysis of the sleep macrostructure is thus of interest, given the potential great savings in terms of time and human resources. An additional advantage is the possibility of providing deterministic (repeatable) diagnostic outcomes, hence contributing to the standardization and quality improvement in the diagnosis. The topic, in fact, is not new, and first related approximations can be traced back to the 1970's [3] [4]. Countless attempts have followed since then and up to now [5] [6] [7] [8] [9] [10] [11] [12] [13] [14], evidencing that the task still represents a challenge, and an open area of research interest. Indeed, despite the promising validation results in some of these works, practical acceptance of these systems among the clinical community remains low. In particular a great challenge faced by these systems has been related to their inability to sustain their announced results in the research lab, and extend them to the practical clinical environment. Some of the main reasons for this contrast can be found in the limited validation procedures, and the overfitting of the resulting methods to the original validation conditions and to the specific tested databases. Ultimately, the consequence is that the resulting systems do not generalize well, and are unable to address the different sources of variability inherent to the sleep staging task.

More recently, several approximations have been appearing based on the use of deep learning, claiming advantages over previous realizations, including improved performance, and the possibility to skip handcrafted feature engineering processes [15] [16] [17] [18] [19] [20] [21] [22] . A common drawback yet remains the limited validation procedures which lack adequate confrontation with the so-called database variability problem [23]. In particular, estimation of a system's performance is usually approached using a subset of independent (testing) data from the whole set available in a reference database. This testing subset, while independent of the training data, remains effectively "local" to the reference database in the sense that it shares the characteristics specific to the common data generation process. However, when considering a multiple-database validation scenario, involving datasets from different external data sources, variability associated with the specific characteristics of each database represents an extra challenge to the machine learning process, even if all databases refer to the same common target task. Indeed, even though the performance of a computer model might have been evaluated "independently" by setting apart a local testing set, in practice, few (if anything at all) can be concluded on the expected generalization performance when the model is presented with data from an additional external dataset. In the case of sleep staging sources of data variability are multiple and include, for example, differences in subject's conditions or physiology, on the processes of signal acquisition and digitalization, including sampling rates, electrode positions, amplification factors, or noise-to-signal ratios, and also important, differences in the expert's interpretation due to human subjectivity and in the training background. Detailed discussion on the topic can be found in a previous work of the authors [23], in which a general performance downgrading trend has been reported among the few works that have attempted validation procedures involving independent external databases.

In this work we describe a new deep learning approach for automatic sleep staging. We carry out its validation by addressing its generalization capabilities on a wide range of sleep staging databases. To better contextualize the database variability problem and the performance of our approach, prediction capabilities for each database are evaluated in the context of both independent local and external generalization scenarios. In the first case, part of each dataset is set aside to be used as independent testing set, while the rest of the data are used for training and parameterization of the machine learning model. This is the classical and most extended schema used in the related literature. On the second scenario (external database validation) the whole dataset is presented *brand-new* to the machine learning model which was parameterized based on data from completely independent database(s). Effectively, by comparing both procedures it is possible to extrapolate the expected performance of the method, regardless of a specific local database used in the tests, and hence reach a better estimation of the real generalization capabilities of the algorithm on the general reference task of sleep staging. In addition, we also examine the suitability of an a novel approach introduced on a previous work [23] based on the use of an ensemble of individual local models, which shows advantages in terms of modelling and learning scalability, to examine its impact on the resulting inter-database generalization performance. Validation results are contextualized in terms of the expected levels of human expert agreement for the same task, and the performance of the current state-of-the-art solutions for automatic sleep scoring.

## 2. MATERIALS AND METHODS

Through the next subsections we describe our proposed deep learning architecture for automatic sleep staging and the corresponding experimental methodology used for its validation. The validation involves analysis and comparison of the performance generalization results on different sleep staging datasets, both on a local and external validation scenario, as described in the previous section.

### 2.1. Neural Network Architecture

Here we describe the general deep learning architecture proposed for the implementation of an automatic sleep staging model. As illustrated in Figure 1, the general architecture is composed of three main processing modules: (i) pre-processing, (ii) Convolutional Neural Network (CNN), and (iii) a Long-Short Term Memory (LSTM).

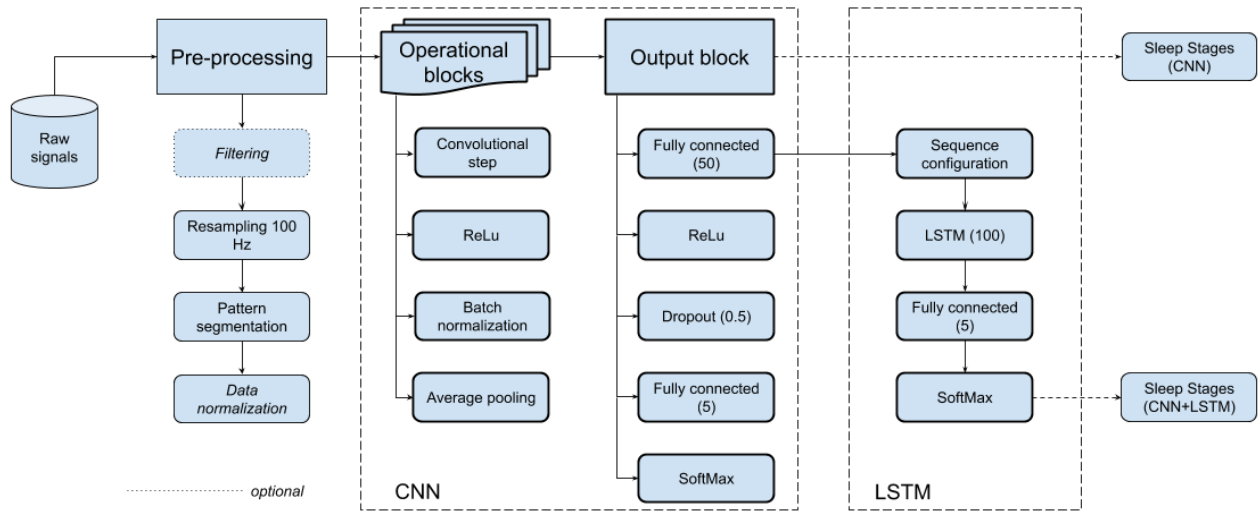

*Figure 1. Preprocessing steps and general CNN-LSTM neural network architecture*

2.1.1. Pre-processing

The preprocessing block is in charge of processing the PSG signals for input homogenization and for (optionally) artifact cancellation. Input signal homogenization is necessary to confer the model the capacity to handle inter-database differences due to the use of different montages and digitalization procedures. Specifically the model receives as input two EEG, the chin EMG, and one EOG channel derivations, which are resampled at 100 Hz, representing a compromise between the limiting the size of the input dimensionality, and the preservation of the necessary signal properties for carrying out the sleep scoring task. Resampling at 100 Hz allows a working frequency up to 50 Hz which captures most of the meaningful EEG, EMG and EOG frequencies. Signals are then segmented using a 30s window following the standard epoch-based scoring procedures [1], resulting on input patterns of size 4x3000 that are fed into the following CNN processing block. Each of these input patterns is subsequently normalized in amplitude using a Gaussian standardization procedure [24]. On a previous study we have shown that this sort of epoch-based normalization procedure considerably improves the generalization capabilities of the network, as it helps dealing with database-specific amplification factors, in contraposition to absence of normalization, or normalization based on long(er)-term data trends [23].

Input signal filtering is left as an optional pre-processing step. As stated before, the main purpose is the removal of noise and signal artifacts, which are patient and database specific, and thus can interfere with the generalization capabilities of the resulting models. Application of the optional filtering step takes place over the original raw signals, i.e. at the original signal frequencies before resampling them at 100 Hz. Experimentation is carried out in this work to study the effects of applying the following pre-processing step on the tested datasets:

-Notch filtering: It is meant to remove the interference caused by the power grid. Notice that the AC frequency differs per country (e.g. 50 Hz in Europe, and 60 Hz in North America) and therefore, depending on the source dataset, mains interference will affect signals at different frequency ranges. Design and implementation of the used digital filter has been described in previous works [25] [26].

-High-pass filter: It is applied to the chin EMG only, and the purpose is to get rid of the DC and low frequency components unrelated to the baseline muscle activity. A first order implementation has been described elsewhere [26]. In this work a cut-off value at 15 Hz has been used for the filter.

-ECG filtering: Applied only in the case that an additional ECG derivation is included in the corresponding montage (see Supplementary Table S1) the filter is used in order to get rid of possible spurious twitches caused by the ECG, affecting the input signals. An adaptive filtering algorithm has been used which has been described in detail in a previous work [25].

2.1.2. CNN block

The CNN block design is an updated version of previous CNN models developed by the authors [19] [23]. As stated before, this block receives input patterns of size 4x3000, representing a 30s epoch window of PSG signals (2xEEG, 1xEMG, and 1xEOG). The block can produce a valid sleep staging output for each input pattern (CNN-only), or act as intermediate processing layer to feed a subsequent LSTM block (CNN-LSTM configuration). Experimentation will be carried out in this work to compare the two possible neural network configurations.

The CNN design is composed of the concatenation of *N* operational blocks. Each operational block *B(k)*, *k = 1...N*, is at the same time composed of four layers, namely (*i*) a 1D convolutional step (kernel size 1x100, preserving the input size with zero padding at edges, stride = 1), followed by (*ii*) ReLu activation [27], (*iii*) batch normalization [28], and (*iv*) an average pool layer (pool dimension 1x2, stride = 2). While the kernel size (1x100) at the convolutional step is maintained through all the *N* operational blocks, the number of filters in *B(k)*, is doubled as with respect to *B(k-1)*. Based on previous experiments [19] [23] the initial number of filters in *B(1)* was set for this work at 8, while the number of operational blocks was fixed as *N = 3*.

Output of the last operational block is fed into a subsequent CNN output block. The first processing layer in the output block is a full-connected step which takes the output from the last operational block and reduces the feature space to an output size of 50. This will be used as the input for the subsequent LSTM processing block when the network is working under the CNN-LSTM configuration. When the network is configured as CNN-only, then four additional processing steps follow. Specifically the 50-length feature vector is filtered through an additional ReLu activation, and then a dropout step with probability 0.5 is applied to improve regularization. Finally a final dense full-connected layer with *softmax* activation is used at the output with size 5, each representing a possible sleep stage assignment (W, N1, N2, N3, or R). The output of the *softmax* is interpreted as the corresponding posterior class probability, with the highest probability determining the final classification decision.

2.1.3. LSTM block

As stated before, when the network follows the CNN-LSTM configuration, the 50-length feature vector is fed into a subsequent LSTM processing block. The inclusion of an additional LSTM layer in the design is meant to provide the resulting network with the capacity of modelling the effect of epoch sequence on the final scoring. Indeed, the medical expert decision on the classification of the current PSG epoch is partially influenced by the sleep state of the preceding and subsequent epochs [1].

The LSTM block is composed of a first sequence configuration layer, a unidirectional LSTM layer [29], and finally, a fully-connected layer followed by *softmax* activation for producing the final output. The sequence configuration step composes the corresponding epoch feature sequence relative to the epoch *k* under evaluation. Specifically given a PSG recording containing *M* epoch intervals, for a given epoch *k, k = 1…M*, the sequence *S(k)* is composed as $[F\left(k - \left\lceil\frac{L-1}{2}\right\rceil\right), F\left(k + 1 - \left\lceil\frac{L-1}{2}\right\rceil\right), ..., F(k-1), F(k), F(k+1), ..., F\left(k - 1 + \left\lceil\frac{L-1}{2}\right\rceil\right), F\left(k + \left\lfloor\frac{L-1}{2}\right\rfloor\right)]$, where $\lceil\ \rceil$ and $\lfloor\ \rfloor$ respectively represent the *ceil* and the *floor* operations, *L* is the length of the sequence, and *F* stands for the corresponding input feature vector, in this case out of the preceding CNN node. For example, if *L = 3*, then the sequence would result as [*F(k-1), F(k), F(K+1)*], and if *L = 4*, then [*F(k-2), F(k-1), F(k), F(k+1)*], and so on. The number of hidden neurons for the LSTM layer was set to 100 in this study.

**2.2. Ensemble of local models**

The intuitive approach to achieve better generalization of a machine learning model is to increment the amount and heterogeneity of the input training data. In the scenario where data from different sources (in our case, different databases) are involved, the former would translate into using data from the all the available datasets. Thereby the amount of training data increases, as well as their heterogeneity, hence boosting the chances of ending up with a true generalist model minimizing the dataset overfitting risk. This approach, however, has its own drawbacks, namely higher memory and computational resources are needed, the resulting model becomes inflexible to data evolving dynamically in time, and a combinatory explosion occurs when finding the best input dataset partition combinations if willing to boost the resulting inter-database generalization capabilities of the model [23].

At this respect a proposal was depicted on a previous work [23] based on the use of an ensemble of local models. Under this approach an independent "local" model is developed for each of the available datasets. For this purpose each dataset is split whereby part of the data are used for training and parameterization of the machine learning model, and the remaining are set aside to be used as independent local testing set. The resulting individual models are then combined by using an ensemble to predict new unseen data from external sources. Specifically, in this work we are assuming that the ensemble output takes place using the majority vote [30] [31]. Such approach shows advantages in the scalability of the design, making it flexible to dynamic evolution of the input datasets, i.e. the ensemble can be easily expanded by adding new local models when new training data is available [23].

In this study we want to check the working hypothesis that by combining "local expert models" by means of an ensemble we can also increase the overall generalization capabilities of the resulting model when predicting external datasets.

**2.3. Experimental design**

The following experimental design is aimed at testing the prediction and generalization capabilities of the general deep learning architecture for automatic sleep staging described in the preceding sections. To adequately address the database variability problem, validation is carried out over a multiple-database scenario in which both, independent local and external database prediction scenarios, are considered.

For this purpose, different clinical sleep scoring datasets, each from an independent database source, were collected. A description of the characteristics of each dataset is provided in the next subsection 2.4. For the purposes of reproducibility, all databases were gathered from public online repositories, with the only exception of our own local sleep center database (not publicly available yet). With no exception, all the databases are digitally encoded using the open EDF(+) format [32] [33].

For each dataset $k$, $k=1...K$, the following experiments are carried out:

Experiment 1:
-Each dataset $k$, is split following an independent training $TR(k)$ and testing $TS(k)$ partition. Let us denote the whole original dataset by $W(k) = TR(k) \cup TS(k)$. A model $M(k)$ is derived by learning from data in $TR(k)$. Notice that a subset of $TR(k)$ –namely the validation subset $VAL(k)$- is used to implement the early stopping criterion during the network's learning process. The "local" generalization performance of the resulting model $M(k)$ is evaluated by assessing the predictability of data contained in $TS(k)$. This is the performance that is usually reported in the literature when data from only one database is used for experimentation.

Experiment 2:
-Each resulting model $M(k)$, is used to predict the reference scorings on each of the complete datasets $W(j)$, $j=1...K$. Effectively, $\forall j / j <> k$, $M(k)$ is predicting unseen data from an external database. Hence, by comparing the results of *Experiment 1* and *Experiment 2*, the effects of varying the database target can be assessed. In effect, for each $M(k)$, the expected local generalization in $TS(k)$ can be compared with the effective inter-database generalization performance among all $W(j)$, $j<>k$. Notice that when $j=k$ the results would be biased since $TR(k) => M(k)$ and $TR(k) \subseteq W(k)$.

Experiment 3:
-Each dataset $W(k)$ is predicted by an ensemble $ENS(k)$ of individual local models $M(j)$, $j=1...K$, $j<>k$. For instance, $ENS(2) = ENS[M(1), M(3), ..., M(K)]$. As in Experiment 2, exclusion of $M(k)$ from $ENS(k)$ aims to keep $W(k)$ completely independent and external to $ENS(k)$. By comparing the results of *Experiment 3* with those of *Experiment 1* and *Experiment 2*, it is possible to assess the effects on the resulting inter-database generalization of the proposed ensemble approach.

In addition, each of the previously described experiments is repeated using different variations of the specific configuration of the deep learning architecture described in Section 2.1. The purpose is to analyze the impact of each configuration on the resulting generalization capabilities of the different models. Specifically, the following variants are tested:

-Using the CCN-only configuration, first the default input segments of 30s (1 epoch, input size 4x3000) are used as described in Section *2.1.2.* for the *CNN block*. The input segments are then modified to form sequences of consecutive epochs with the aim of implementing the effect of epoch sequence learning. Different sequence lengths $L = \{3,5,7\}$ are investigated at this respect. Gaussian normalization takes place in this case over the whole 4x(3000L) resulting input patterns. This approach to implement epoch sequence learning using a CCN-only configuration will be afterwards compared to the results achieved using the described full CNN-LSTM design.

-Using CNN-LSTM configuration, as described in Section 2.1.3, the sequence length parameter is set equivalently to $L = \{3,5,7\}$, i.e. using as input the 50-length feature vector of the preceding CNN output block. As stated before, the resulting models will be compared against the respective sequence learning implementations using the CNN-only configuration.

-Finally, in order to test the effects of the optional signal preprocessing filtering step, each of the previous described experiments are performed, respectively, with and without applying the filtering pipeline described in Section 2.1.1.

Thus, for each of the datasets included in our experimentation, a total of 14 different individual local models are developed, based on the data contained on each respective dataset. For identification, the following nomenclature is used: *CNN_1, CNN_3, CNN_5, CNN_7, CNN_F_1, CNN_F_3, CNN_F_5, CNN_F_7, CNN_LSTM_3, CNN_LSTM_5, CNN_LSTM_7, CNN_LSTM_F_3, CNN_LSTM_F_5, CNN_LSTM_F_7,* where the subscript *F* denotes the use of the pre-processing filtering step, and the suffix number indicates the corresponding number of sequence epochs used (value of the *L* parameter).

For homogenization purposes, the same training configuration is applied in the development of above mentioned learning models for each dataset. In this respect the stochastic gradient descent approach is used to guide the weight's update, with the cross-entropy loss as the target cost function [24]. Each dataset is partitioned using 80% of data for training (TR), using the remaining 20% as independent local testing set (TS). A validation subset (VAL) is arranged by successively splitting 20% of the available training data apart. The validation set is used as reference to implement the early stopping mechanism to avoid overfitting to training data. The stopping criterion takes as reference the validation loss, which is evaluated 5 times per training epoch. A patience of 10 is established thereby stopping training when the validation loss has not been further improved after the whole training dataset is presented two times. The number of patterns within each training epoch (internal training batch) is set to 100 patterns, imposed by the available hardware resources relative to the size of the tested networks. The maximum number of training epochs is set to 30, and the initial learning rate to 0.001. The learning rate is decreased by a factor of 10 every 10 training epochs (thus $10^{-4}$, $10^{-5}$, up to a minimum of $10^{-6}$). The same random initialization seed is used on each experiment to exclude variability due to initialization conditions, hence enabling deterministic training processes. This is important to assess the influence of the different tested architecture variants, as described before, and to make fair comparisons among the different resulting models and datasets.

Performance evaluation on each experiment is carried out taking the Cohen's kappa index (κ) as the reference validation score. Cohen's kappa is preferred over other widespread validation metrics (e.g. classification error, sensitivity/specificity, or $F_1$-score) because it accounts for agreement due to chance, and it shows robustness in the presence of various class distributions [34]. This is an important property to allow performance comparison among differently distributed datasets. Remarkably, Cohen's kappa has been widely reported as the reference validation metric among many studies analyzing human inter-rater variability in the context of sleep scoring (see Section 4.5 for more reference).

## 2.4. Datasets

A set of heterogeneous and independent PSG datasets was used as testing benchmark during the course of our experiments. In order to enhance reproducibility, the list mostly includes public well-known datasets, with the only exception of a subset of our own in-house patient database (HMC dataset, described below, not yet publicly available). The same database benchmark has been used to test a previous approach of the authors [23], which enables setting up a performance baseline for direct comparison of the results. Subsequently, an overview of each integrating dataset is given. Extended description, including specifications of the corresponding signal montages for each dataset, can also be found in the Supplementary Table S1.

Haaglanden Medisch Centrum Sleep Center Database (HMC)
This dataset includes a total of 159 recordings gathered from the sleep center database of the Haaglanden Medisch Centrum (The Netherlands) during April 2018. Patient recordings were randomly selected and include a heterogeneous population which was referred for PSG examination on the context of different sleep disorders. The recordings were acquired in the course of common clinical practice, and thus did not subject people to any other treatment nor prescribed any additional behavior outside of the usual clinical procedures. Data were anonymized avoiding any possibility of individual patient identification. Ethical approval from the Zuid-West Holland committee for using this dataset was granted under identification code METC-19-065. This is the only dataset that cannot be found publicly available online.

St. Vicent's Hospital / University College Dublin Sleep Apnea Database (Dublin)
This dataset contains 25 full overnight polysomnograms from adult subjects with suspected sleep-disordered breathing. Subjects were originally randomly selected over a 6-month period (September 02 to February 03) from patients referred to the Sleep Disorders Clinic at St Vincent's University Hospital, Dublin, for possible diagnosis of obstructive sleep apnea, central sleep apnea or primary snoring. The dataset is available online on the PhysioNet website [35].

Sleep Health Heart Study (SHHS)
The Sleep Heart Health Study (SHHS) is a multi-center cohort study implemented by the National Heart Lung & Blood Institute to determine the cardiovascular and other consequences of sleep-disordered breathing. The database is available online upon permission at the National Sleep Research Resource (NSRR) [36] [37]. More information about the rationale, design, and protocol of the SHHS study can be found in the dedicated NSRR section [37] and in the literature [38] [39]. For this study a random subset of 100 PSG recordings were selected from the SHHS-2 study.

Sleep Telemetry Study (Telemetry)
This dataset contains 44 whole-night PSGs obtained in a 1994 study of temazepam effects on sleep in 22 caucasian males and females without other medication. Subjects had mild difficulty falling asleep but were otherwise healthy. The PSGs were recorded in the hospital during two nights, one of which was after temazepam intake, and the other of which was after placebo intake. More details on the subjects and the recording conditions are further described in the works of Kemp et al. [40] [41]. The dataset is fully available at the PhysioNet website as part of the more extensive Sleep-EDF database [42].

DREAMS Subject database (DREAMS)
The DREAMS dataset is composed of 20 whole-night PSG recordings from healthy subjects. It was collected during the DREAMS project, to tune, train, and test automatic sleep staging algorithms [43]. The dataset is available online granted by University of MONS - TCTS Laboratory (Stéphanie Devuyst, Thierry Dutoit) and Université Libre de Bruxelles - CHU de Charleroi Sleep Laboratory (Myriam Kerkhofs) under terms of the Attribution-NonCommercial-NoDerivs 3.0 Unported (CC BY-NC-ND 3.0) [44].

ISRUC-SLEEP Dataset (ISRUC)
This dataset is composed of 100 PSGs from adult subjects with evidence of having sleep disorders. PSG recordings were originally selected from the Sleep Medicine Centre of the Hospital of Coimbra University (CHUC) database during the period 2009–2013. More details about the rationale and the design of the database can be found in Khalighi et al. [45]. The database is publicly accessible online [46].

It is worth to mention that no exclusion criteria were applied a posteriori on any of the involved datasets. Thus, all the recordings integrating the respective original selections were included for validation purposes. The underlying motivation is to assess the reliability of the resulting models on the most realistic situation, including the most general and heterogeneous patient phenotype possible. The AASM scoring standard was used as reference for the output class labels. Hence, when the original dataset was scored using the R&K method, NREM stages 3 and 4 were merged into the corresponding N3, according to the AASM guidelines [1]. Notice as well that the specific signal montages can differ across the different source databases. As introduced in Section 2.1.1., our deep learning model assumes as input two channels of EEG, one submental EMG, and one EOG derivation. If available in the corresponding montage, and only when the optional filtering preprocessing step is applied, an additional ECG derivation can be used for the purposes of artifact removal on the input signals. The additional ECG channel, however, is never used as a direct input to the learning model. Supplementary Table S1 describes the specific selected derivations according to the available set of channels as well as the main characteristics of each dataset.

## 3. RESULTS

The following tables contain the results of the experiments described in Section 2.3.

Table 1 shows the results of Experiment 1, where each of the learning models is trained and evaluated using data from its respective local testing dataset.

*Table 1. Performance results of each individual model on the local validation scenario. For each dataset results are given with respect to the corresponding training (TR), validation (VAL) and testing (TS) dataset partitions. The number of effective training iterations is indicated in the third column. Rows within each dataset correspond to the different tested neural network configurations. Results are reported in terms of kappa index with respect to the corresponding clinical scorings of reference for each dataset.*

| Local dataset | Model configuration | Training iterations | TR | VAL | TS |
| --- | --- | --- | --- | --- | --- |
| HMC | CNN_1 | 15 | 0.79 | 0.73 | 0.74 |
| | CNN_3 | 7 | 0.83 | 0.72 | 0.71 |
| | CNN_5 | 7 | 0.87 | 0.71 | 0.70 |
| | CNN_7 | 5 | 0.83 | 0.69 | 0.69 |
| | CNN_LSTM_3 | 7 | 0.81 | 0.78 | 0.78 |
| | CNN_LSTM_5 | 17 | 0.84 | 0.79 | 0.79 |
| | CNN_LSTM_7 | 27 | 0.83 | 0.77 | 0.77 |
| | CNN_F_1 | 14 | 0.78 | 0.73 | 0.74 |
| | CNN_F_3 | 8 | 0.84 | 0.71 | 0.71 |
| | CNN_F_5 | 6 | 0.84 | 0.70 | 0.70 |
| | CNN_F_7 | 5 | 0.85 | 0.69 | 0.69 |
| | CNN_LSTM_F_3 | 7 | 0.79 | 0.77 | 0.77 |
| | CNN_LSTM_F_5 | 8 | 0.77 | 0.75 | 0.75 |
| | CNN_LSTM_F_7 | 10 | 0.76 | 0.74 | 0.74 |
| Dublin | CNN_1 | 10 | 0.76 | 0.68 | 0.68 |
| | CNN_3 | 7 | 0.85 | 0.66 | 0.66 |
| | CNN_5 | 6 | 0.89 | 0.62 | 0.64 |
| | CNN_7 | 7 | 0.89 | 0.65 | 0.67 |
| | CNN_LSTM_3 | 8 | 0.82 | 0.76 | 0.77 |
| | CNN_LSTM_5 | 9 | 0.84 | 0.78 | 0.79 |
| | CNN_LSTM_7 | 9 | 0.84 | 0.77 | 0.77 |
| | CNN_F_1 | 14 | 0.77 | 0.65 | 0.65 |
| | CNN_F_3 | 8 | 0.83 | 0.65 | 0.64 |
| | CNN_F_5 | 8 | 0.88 | 0.60 | 0.61 |
| | CNN_F_7 | 13 | 0.90 | 0.67 | 0.66 |
| | CNN_LSTM_F_3 | 8 | 0.81 | 0.76 | 0.77 |
| | CNN_LSTM_F_5 | 8 | 0.82 | 0.78 | 0.79 |
| | CNN_LSTM_F_7 | 9 | 0.84 | 0.77 | 0.78 |
| SHHS | CNN_1 | 9 | 0.80 | 0.76 | 0.75 |
| | CNN_3 | 7 | 0.89 | 0.79 | 0.79 |
| | CNN_5 | 7 | 0.95 | 0.78 | 0.79 |
| | CNN_7 | 6 | 0.92 | 0.76 | 0.76 |
| | CNN_LSTM_3 | 17 | 0.87 | 0.83 | 0.84 |
| | CNN_LSTM_5 | 18 | 0.86 | 0.83 | 0.82 |
| | CNN_LSTM_7 | 7 | 0.79 | 0.78 | 0.77 |
| | CNN_F_1 | 9 | 0.80 | 0.77 | 0.76 |
| | CNN_F_3 | 8 | 0.89 | 0.79 | 0.79 |
| | CNN_F_5 | 5 | 0.90 | 0.78 | 0.78 |
| | CNN_F_7 | 6 | 0.94 | 0.78 | 0.77 |
| | CNN_LSTM_F_3 | 10 | 0.85 | 0.82 | 0.83 |
| | CNN_LSTM_F_5 | 18 | 0.86 | 0.83 | 0.82 |
| | CNN_LSTM_F_7 | 5 | 0.80 | 0.79 | 0.79 |
| Telemetry | CNN_1 | 14 | 0.81 | 0.76 | 0.76 |
| | CNN_3 | 10 | 0.88 | 0.73 | 0.75 |
| | CNN_5 | 8 | 0.90 | 0.73 | 0.72 |
| | CNN_7 | 6 | 0.88 | 0.72 | 0.70 |
| | CNN_LSTM_3 | 10 | 0.85 | 0.80 | 0.81 |
| | CNN_LSTM_5 | 9 | 0.85 | 0.79 | 0.80 |
| | CNN_LSTM_7 | 8 | 0.84 | 0.80 | 0.80 |
| | CNN_F_1 | 14 | 0.82 | 0.77 | 0.77 |
| | CNN_F_3 | 8 | 0.87 | 0.71 | 0.71 |
| | CNN_F_5 | 9 | 0.88 | 0.73 | 0.73 |
| | CNN_F_7 | 9 | 0.91 | 0.73 | 0.71 |
| | CNN_LSTM_F_3 | 10 | 0.84 | 0.81 | 0.81 |
| | CNN_LSTM_F_5 | 15 | 0.89 | 0.82 | 0.83 |
| | CNN_LSTM_F_7 | 16 | 0.89 | 0.82 | 0.82 |
| DREAMS | CNN_1 | 9 | 0.81 | 0.75 | 0.76 |
| | CNN_3 | 8 | 0.92 | 0.76 | 0.76 |
| | CNN_5 | 7 | 0.91 | 0.75 | 0.75 |
| | CNN_7 | 5 | 0.85 | 0.73 | 0.73 |
| | CNN_LSTM_3 | 17 | 0.88 | 0.84 | 0.83 |
| | CNN_LSTM_5 | 20 | 0.90 | 0.83 | 0.83 |
| | CNN_LSTM_7 | 5 | 0.81 | 0.78 | 0.78 |
| | CNN_F_1 | 9 | 0.82 | 0.76 | 0.77 |
| | CNN_F_3 | 7 | 0.91 | 0.77 | 0.78 |
| | CNN_F_5 | 7 | 0.92 | 0.74 | 0.75 |
| | CNN_F_7 | 6 | 0.88 | 0.73 | 0.72 |
| | CNN_LSTM_F_3 | 20 | 0.89 | 0.84 | 0.83 |

| | | | | | | |
|---|---|---|---|---|---|---|
| | CNN_LSTM_F_5 | 28 | 0.90 | 0.84 | 0.84 | |
| | CNN_LSTM_F_7 | 10 | 0.85 | 0.81 | 0.80 | |
| ISRUC | CNN_1 | 17 | 0.81 | 0.77 | 0.76 | |
| | CNN_3 | 6 | 0.83 | 0.74 | 0.75 | |
| | CNN_5 | 7 | 0.90 | 0.73 | 0.73 | |
| | CNN_7 | 6 | 0.86 | 0.72 | 0.73 | |
| | CNN_LSTM_3 | 10 | 0.81 | 0.80 | 0.80 | |
| | CNN_LSTM_5 | 10 | 0.80 | 0.78 | 0.78 | |
| | CNN_LSTM_7 | 6 | 0.75 | 0.75 | 0.75 | |
| | CNN_F_1 | 9 | 0.79 | 0.76 | 0.75 | |
| | CNN_F_3 | 7 | 0.84 | 0.75 | 0.75 | |
| | CNN_F_5 | 7 | 0.90 | 0.73 | 0.73 | |
| | CNN_F_7 | 6 | 0.86 | 0.71 | 0.72 | |
| | CNN_LSTM_F_3 | 10 | 0.81 | 0.79 | 0.79 | |
| | CNN_LSTM_F_5 | 9 | 0.78 | 0.77 | 0.76 | |
| | CNN_LSTM_F_7 | 9 | 0.75 | 0.74 | 0.74 | |

Subsequent Table 2 shows the results of the second experiment in which the resulting individual local models have been used to predict the reference scorings on each of the complete datasets. Results in Table 2 therefore involve performance evaluations of the models using an external validation setting, with the only exception of the main diagonal. The main diagonal in Table 2 represents the situation in which *M(k)* is used to predict *W(k)*, resulting in a biased prediction since $TR(k) => M(k)$ and $TR(k) \subseteq W(k)$. Regardless, these results have been kept in Table 2 for reference.

*Table 2. Performance results of the individual local models on the external validation scenario. The notation M(X) is used to indicate that the model was trained based on data on the dataset X. Rows within each dataset correspond to the different tested neural network configurations. The main diagonal (in greyed background) shows the results when the model is predicting its own complete local dataset (biased prediction). Results are reported in terms of kappa index with respect to the corresponding clinical scorings of reference for each dataset.*

| | | Individual local models | | | | | |
|---|---|---|---|---|---|---|---|
| Predicted dataset | Model configuration | M(HMC) | M(Dublin) | M(SHHS) | M(Telemetry) | M(DREAMS) | M(ISRUC) |
| HMC | CNN_1 | 0.77 | 0.51 | 0.56 | 0.53 | 0.52 | 0.60 |
| | CNN_3 | 0.79 | 0.46 | 0.60 | 0.42 | 0.47 | 0.56 |
| | CNN_5 | 0.81 | 0.37 | 0.57 | 0.39 | 0.44 | 0.54 |
| | CNN_7 | 0.78 | 0.40 | 0.50 | 0.34 | 0.43 | 0.55 |
| | CNN_LSTM_3 | 0.80 | 0.54 | 0.58 | 0.51 | 0.50 | 0.61 |
| | CNN_LSTM_5 | 0.82 | 0.53 | 0.60 | 0.50 | 0.49 | 0.62 |
| | CNN_LSTM_7 | 0.81 | 0.52 | 0.59 | 0.52 | 0.51 | 0.61 |
| | CNN_F_1 | 0.76 | 0.39 | 0.58 | 0.49 | 0.56 | 0.62 |
| | CNN_F_3 | 0.79 | 0.37 | 0.60 | 0.48 | 0.48 | 0.63 |
| | CNN_F_5 | 0.79 | 0.35 | 0.57 | 0.42 | 0.46 | 0.61 |
| | CNN_F_7 | 0.79 | 0.35 | 0.54 | 0.37 | 0.50 | 0.58 |
| | CNN_LSTM_F_3 | 0.78 | 0.38 | 0.62 | 0.45 | 0.55 | 0.64 |
| | CNN_LSTM_F_5 | 0.76 | 0.37 | 0.63 | 0.48 | 0.55 | 0.65 |
| | CNN_LSTM_F_7 | 0.75 | 0.34 | 0.62 | 0.47 | 0.55 | 0.62 |
| Dublin | CNN_1 | 0.53 | 0.73 | 0.44 | 0.41 | 0.53 | 0.51 |
| | CNN_3 | 0.57 | 0.78 | 0.50 | 0.34 | 0.49 | 0.57 |
| | CNN_5 | 0.52 | 0.79 | 0.51 | 0.32 | 0.51 | 0.59 |
| | CNN_7 | 0.53 | 0.81 | 0.39 | 0.31 | 0.49 | 0.57 |
| | CNN_LSTM_3 | 0.54 | 0.80 | 0.48 | 0.38 | 0.58 | 0.55 |
| | CNN_LSTM_5 | 0.54 | 0.82 | 0.50 | 0.39 | 0.58 | 0.55 |
| | CNN_LSTM_7 | 0.50 | 0.81 | 0.50 | 0.42 | 0.57 | 0.55 |
| | CNN_F_1 | 0.20 | 0.73 | 0.13 | 0.03 | 0.01 | 0.07 |
| | CNN_F_3 | 0.15 | 0.77 | 0.10 | 0.01 | 0.02 | 0.05 |
| | CNN_F_5 | 0.04 | 0.78 | 0.02 | 0.01 | 0.03 | 0.04 |
| | CNN_F_7 | 0.04 | 0.81 | 0.13 | 0.01 | 0.02 | 0.14 |
| | CNN_LSTM_F_3 | 0.24 | 0.80 | 0.07 | 0.02 | 0.01 | 0.05 |
| | CNN_LSTM_F_5 | 0.22 | 0.81 | 0.07 | 0.01 | 0.01 | 0.05 |
| | CNN_LSTM_F_7 | 0.17 | 0.81 | 0.06 | 0.01 | 0.01 | 0.04 |
| SHHS | CNN_1 | 0.57 | 0.50 | 0.78 | 0.42 | 0.59 | 0.64 |
| | CNN_3 | 0.59 | 0.52 | 0.86 | 0.30 | 0.60 | 0.63 |
| | CNN_5 | 0.55 | 0.42 | 0.89 | 0.27 | 0.60 | 0.63 |
| | CNN_7 | 0.61 | 0.40 | 0.86 | 0.26 | 0.60 | 0.65 |
| | CNN_LSTM_3 | 0.54 | 0.57 | 0.86 | 0.42 | 0.54 | 0.62 |
| | CNN_LSTM_5 | 0.50 | 0.56 | 0.85 | 0.46 | 0.52 | 0.67 |
| | CNN_LSTM_7 | 0.47 | 0.53 | 0.78 | 0.46 | 0.56 | 0.66 |
| | CNN_F_1 | 0.68 | 0.35 | 0.78 | 0.43 | 0.60 | 0.65 |
| | CNN_F_3 | 0.53 | 0.29 | 0.85 | 0.38 | 0.59 | 0.65 |
| | CNN_F_5 | 0.52 | 0.32 | 0.86 | 0.39 | 0.58 | 0.68 |
| | CNN_F_7 | 0.52 | 0.28 | 0.88 | 0.31 | 0.63 | 0.67 |

| Dataset | Model | | | | | | |
|---|---|---|---|---|---|---|---|
| | CNN_LSTM_F_3 | 0.68 | 0.29 | 0.84 | 0.40 | 0.57 | 0.63 |
| | CNN_LSTM_F_5 | 0.66 | 0.31 | 0.85 | 0.39 | 0.53 | 0.65 |
| | CNN_LSTM_F_7 | 0.67 | 0.22 | 0.79 | 0.41 | 0.57 | 0.62 |
| Telemetry | CNN_1 | 0.67 | 0.53 | 0.51 | 0.79 | 0.48 | 0.63 |
| | CNN_3 | 0.60 | 0.42 | 0.55 | 0.81 | 0.43 | 0.53 |
| | CNN_5 | 0.60 | 0.43 | 0.54 | 0.83 | 0.48 | 0.57 |
| | CNN_7 | 0.50 | 0.45 | 0.38 | 0.82 | 0.45 | 0.52 |
| | CNN_LSTM_3 | 0.68 | 0.59 | 0.49 | 0.83 | 0.45 | 0.62 |
| | CNN_LSTM_5 | 0.69 | 0.59 | 0.50 | 0.83 | 0.43 | 0.64 |
| | CNN_LSTM_7 | 0.67 | 0.61 | 0.50 | 0.82 | 0.48 | 0.65 |
| | CNN_F_1 | 0.70 | 0.39 | 0.61 | 0.80 | 0.44 | 0.61 |
| | CNN_F_3 | 0.67 | 0.46 | 0.57 | 0.83 | 0.46 | 0.62 |
| | CNN_F_5 | 0.63 | 0.43 | 0.41 | 0.83 | 0.42 | 0.55 |
| | CNN_F_7 | 0.64 | 0.44 | 0.33 | 0.84 | 0.46 | 0.54 |
| | CNN_LSTM_F_3 | 0.72 | 0.44 | 0.60 | 0.83 | 0.43 | 0.60 |
| | CNN_LSTM_F_5 | 0.71 | 0.48 | 0.62 | 0.87 | 0.44 | 0.63 |
| | CNN_LSTM_F_7 | 0.68 | 0.44 | 0.60 | 0.86 | 0.44 | 0.61 |
| DREAMS | CNN_1 | 0.50 | 0.56 | 0.58 | 0.34 | 0.79 | 0.71 |
| | CNN_3 | 0.46 | 0.52 | 0.59 | 0.34 | 0.86 | 0.71 |
| | CNN_5 | 0.42 | 0.33 | 0.36 | 0.31 | 0.85 | 0.68 |
| | CNN_7 | 0.61 | 0.51 | 0.58 | 0.27 | 0.81 | 0.67 |
| | CNN_LSTM_3 | 0.50 | 0.56 | 0.54 | 0.43 | 0.87 | 0.73 |
| | CNN_LSTM_5 | 0.41 | 0.56 | 0.54 | 0.47 | 0.87 | 0.75 |
| | CNN_LSTM_7 | 0.45 | 0.50 | 0.51 | 0.42 | 0.80 | 0.74 |
| | CNN_F_1 | 0.52 | 0.39 | 0.66 | 0.42 | 0.80 | 0.70 |
| | CNN_F_3 | 0.59 | 0.25 | 0.56 | 0.46 | 0.86 | 0.67 |
| | CNN_F_5 | 0.55 | 0.36 | 0.55 | 0.40 | 0.86 | 0.65 |
| | CNN_F_7 | 0.52 | 0.32 | 0.58 | 0.35 | 0.82 | 0.71 |
| | CNN_LSTM_F_3 | 0.53 | 0.32 | 0.60 | 0.43 | 0.87 | 0.71 |
| | CNN_LSTM_F_5 | 0.55 | 0.31 | 0.63 | 0.46 | 0.88 | 0.72 |
| | CNN_LSTM_F_7 | 0.51 | 0.16 | 0.60 | 0.43 | 0.83 | 0.71 |
| ISRUC | CNN_1 | 0.56 | 0.57 | 0.60 | 0.29 | 0.63 | 0.79 |
| | CNN_3 | 0.57 | 0.54 | 0.64 | 0.29 | 0.56 | 0.80 |
| | CNN_5 | 0.51 | 0.46 | 0.63 | 0.26 | 0.57 | 0.84 |
| | CNN_7 | 0.57 | 0.48 | 0.54 | 0.24 | 0.52 | 0.81 |
| | CNN_LSTM_3 | 0.54 | 0.55 | 0.65 | 0.36 | 0.61 | 0.81 |
| | CNN_LSTM_5 | 0.51 | 0.55 | 0.66 | 0.42 | 0.58 | 0.79 |
| | CNN_LSTM_7 | 0.43 | 0.53 | 0.60 | 0.38 | 0.60 | 0.75 |
| | CNN_F_1 | 0.68 | 0.42 | 0.63 | 0.42 | 0.65 | 0.77 |
| | CNN_F_3 | 0.59 | 0.35 | 0.65 | 0.41 | 0.61 | 0.81 |
| | CNN_F_5 | 0.57 | 0.41 | 0.66 | 0.37 | 0.57 | 0.84 |
| | CNN_F_7 | 0.55 | 0.38 | 0.62 | 0.35 | 0.56 | 0.81 |
| | CNN_LSTM_F_3 | 0.68 | 0.41 | 0.68 | 0.40 | 0.63 | 0.80 |
| | CNN_LSTM_F_5 | 0.67 | 0.43 | 0.69 | 0.44 | 0.61 | 0.78 |
| | CNN_LSTM_F_7 | 0.66 | 0.29 | 0.66 | 0.43 | 0.61 | 0.75 |

Results regarding the third experiment (ensemble predictions) are shown in Table 3, which are compared to the reference predictions of the individual local models, both in the local and external validation scenarios. The third column in Table 3 shows the reference local predictions achieved by the models in their respective testing sets (last column of Table 1). Subsequently, the fourth column shows the corresponding ranges of the inter-database external predictions as derived from data in Table 2. These ranges exclude data from the main diagonal of Table 2, i.e. for dataset *k*, performance of *M(k)* is excluded, hence regarding performance when the individual models are presented with the external datasets exclusively. The resulting average performance is shown in the fifth column. Finally, the last column of Table 3 shows the corresponding performance when the ensemble model is used for predicting the corresponding dataset. Similarly, *ENS(k)* excludes *M(k)* from the ensemble, e.g. for HMC, the derived predictions result from *ENS[M(Dublin), M(SHHS), M(Telemetry), M(DREAMS), M(ISRUC)]*, and so for.

Table 3. Performance comparison between individual models and ensemble approach in the local and external validation scenarios. Rows within each dataset correspond to the different tested neural network configurations. Results are reported in terms of kappa index with respect to the corresponding clinical scorings of reference.

| Predicted dataset | Model configuration | Individual local models | | | Ensemble |
|---|---|---|---|---|---|
| | | Local performance | External performance | | External performance |
| | | | Range | Average | |
| HMC | CNN_1 | 0.74 | 0.51 – 0.60 | 0.54 | 0.61 |
| | CNN_3 | 0.71 | 0.42 – 0.60 | 0.50 | 0.58 |
| | CNN_5 | 0.70 | 0.37 – 0.57 | 0.46 | 0.55 |
| | CNN_7 | 0.69 | 0.34 – 0.55 | 0.44 | 0.53 |
| | CNN_LSTM_3 | 0.78 | 0.50 – 0.61 | 0.55 | 0.62 |
| | CNN_LSTM_5 | 0.79 | 0.49 – 0.62 | 0.55 | 0.63 |
| | CNN_LSTM_7 | 0.77 | 0.51 – 0.61 | 0.55 | 0.62 |
| | CNN_F_1 | 0.74 | 0.39 – 0.62 | 0.53 | 0.61 |
| | CNN_F_3 | 0.71 | 0.37 – 0.63 | 0.51 | 0.60 |
| | CNN_F_5 | 0.70 | 0.35 – 0.61 | 0.48 | 0.58 |
| | CNN_F_7 | 0.69 | 0.35 – 0.58 | 0.47 | 0.56 |
| | CNN_LSTM_F_3 | 0.77 | 0.38 – 0.64 | 0.53 | 0.62 |
| | CNN_LSTM_F_5 | 0.75 | 0.37 – 0.65 | 0.54 | 0.64 |
| | CNN_LSTM_F_7 | 0.74 | 0.34 – 0.62 | 0.52 | 0.63 |
| Dublin | CNN_1 | 0.68 | 0.41 – 0.53 | 0.49 | 0.60 |
| | CNN_3 | 0.66 | 0.34 – 0.57 | 0.49 | 0.62 |
| | CNN_5 | 0.64 | 0.32 – 0.59 | 0.49 | 0.60 |
| | CNN_7 | 0.67 | 0.31 – 0.57 | 0.46 | 0.59 |
| | CNN_LSTM_3 | 0.77 | 0.38 – 0.58 | 0.51 | 0.63 |
| | CNN_LSTM_5 | 0.79 | 0.39 – 0.58 | 0.51 | 0.63 |
| | CNN_LSTM_7 | 0.77 | 0.42 – 0.57 | 0.51 | 0.62 |
| | CNN_F_1 | 0.65 | 0.01 – 0.20 | 0.09 | 0.08 |
| | CNN_F_3 | 0.64 | 0.01 – 0.15 | 0.07 | 0.04 |
| | CNN_F_5 | 0.61 | 0.01 – 0.04 | 0.03 | 0.01 |
| | CNN_F_7 | 0.66 | 0.01 – 0.14 | 0.07 | 0.03 |
| | CNN_LSTM_F_3 | 0.77 | 0.01 – 0.24 | 0.08 | 0.06 |
| | CNN_LSTM_F_5 | 0.79 | 0.01 – 0.22 | 0.07 | 0.05 |
| | CNN_LSTM_F_7 | 0.78 | 0.01 – 0.17 | 0.06 | 0.04 |
| SHHS | CNN_1 | 0.75 | 0.42 – 0.64 | 0.54 | 0.62 |
| | CNN_3 | 0.79 | 0.30 – 0.63 | 0.53 | 0.65 |
| | CNN_5 | 0.79 | 0.27 – 0.63 | 0.49 | 0.61 |
| | CNN_7 | 0.76 | 0.26 – 0.65 | 0.50 | 0.65 |
| | CNN_LSTM_3 | 0.84 | 0.42 – 0.62 | 0.54 | 0.62 |
| | CNN_LSTM_5 | 0.82 | 0.46 – 0.67 | 0.54 | 0.61 |
| | CNN_LSTM_7 | 0.77 | 0.46 – 0.66 | 0.54 | 0.61 |
| | CNN_F_1 | 0.76 | 0.35 – 0.68 | 0.54 | 0.66 |
| | CNN_F_3 | 0.79 | 0.29 – 0.65 | 0.49 | 0.62 |
| | CNN_F_5 | 0.78 | 0.32 – 0.68 | 0.50 | 0.62 |
| | CNN_F_7 | 0.77 | 0.28 – 0.67 | 0.48 | 0.62 |
| | CNN_LSTM_F_3 | 0.83 | 0.29 – 0.68 | 0.52 | 0.62 |
| | CNN_LSTM_F_5 | 0.82 | 0.31 – 0.66 | 0.51 | 0.62 |
| | CNN_LSTM_F_7 | 0.79 | 0.22 – 0.67 | 0.50 | 0.62 |
| Telemetry | CNN_1 | 0.76 | 0.48 – 0.67 | 0.56 | 0.67 |
| | CNN_3 | 0.75 | 0.42 – 0.60 | 0.51 | 0.61 |
| | CNN_5 | 0.72 | 0.43 – 0.60 | 0.53 | 0.62 |
| | CNN_7 | 0.70 | 0.38 – 0.52 | 0.46 | 0.58 |
| | CNN_LSTM_3 | 0.81 | 0.45 – 0.68 | 0.57 | 0.67 |
| | CNN_LSTM_5 | 0.80 | 0.43 – 0.69 | 0.57 | 0.69 |
| | CNN_LSTM_7 | 0.80 | 0.48 – 0.67 | 0.58 | 0.68 |
| | CNN_F_1 | 0.77 | 0.39 – 0.70 | 0.55 | 0.66 |
| | CNN_F_3 | 0.71 | 0.46 – 0.67 | 0.56 | 0.66 |
| | CNN_F_5 | 0.73 | 0.41- 0.63 | 0.49 | 0.62 |
| | CNN_F_7 | 0.71 | 0.33 – 0.64 | 0.48 | 0.63 |
| | CNN_LSTM_F_3 | 0.81 | 0.43 – 0.72 | 0.56 | 0.69 |
| | CNN_LSTM_F_5 | 0.83 | 0.44 – 0.71 | 0.58 | 0.70 |
| | CNN_LSTM_F_7 | 0.82 | 0.44 – 0.68 | 0.56 | 0.68 |
| DREAMS | CNN_1 | 0.76 | 0.34 – 0.71 | 0.54 | 0.61 |
| | CNN_3 | 0.76 | 0.34 – 0.71 | 0.52 | 0.61 |
| | CNN_5 | 0.75 | 0.31 – 0.68 | 0.42 | 0.56 |
| | CNN_7 | 0.73 | 0.27 – 0.67 | 0.53 | 0.63 |
| | CNN_LSTM_3 | 0.83 | 0.43 – 0.73 | 0.55 | 0.61 |
| | CNN_LSTM_5 | 0.83 | 0.41 – 0.75 | 0.55 | 0.59 |
| | CNN_LSTM_7 | 0.78 | 0.42 – 0.74 | 0.52 | 0.58 |
| | CNN_F_1 | 0.77 | 0.39 – 0.70 | 0.54 | 0.62 |
| | CNN_F_3 | 0.78 | 0.25 – 0.67 | 0.51 | 0.66 |
| | CNN_F_5 | 0.75 | 0.36 – 0.65 | 0.50 | 0.64 |
| | CNN_F_7 | 0.72 | 0.32 – 0.71 | 0.50 | 0.63 |
| | CNN_LSTM_F_3 | 0.83 | 0.32 – 0.71 | 0.52 | 0.61 |
| | CNN_LSTM_F_5 | 0.84 | 0.31 – 0.72 | 0.54 | 0.64 |
| | CNN_LSTM_F_7 | 0.80 | 0.16 – 0.71 | 0.48 | 0.59 |
| ISRUC | CNN_1 | 0.76 | 0.29 – 0.63 | 0.53 | 0.59 |
| | CNN_3 | 0.75 | 0.29 – 0.64 | 0.52 | 0.61 |
| | CNN_5 | 0.73 | 0.26 – 0.63 | 0.48 | 0.58 |
| | CNN_7 | 0.73 | 0.24 – 0.57 | 0.47 | 0.56 |
| | CNN_LSTM_3 | 0.80 | 0.36 – 0.65 | 0.54 | 0.60 |
| | CNN_LSTM_5 | 0.78 | 0.42 – 0.66 | 0.54 | 0.61 |
| | CNN_LSTM_7 | 0.75 | 0.38 – 0.60 | 0.51 | 0.57 |
| | CNN_F_1 | 0.75 | 0.42 – 0.68 | 0.56 | 0.64 |
| | CNN_F_3 | 0.75 | 0.35 – 0.65 | 0.52 | 0.63 |

| | | | | |
|---|---|---|---|---|
| CNN_F_5 | 0.73 | 0.37 – 0.66 | 0.51 | 0.60 |
| CNN_F_7 | 0.72 | 0.35 – 0.62 | 0.49 | 0.58 |
| CNN_LSTM_F_3 | 0.79 | 0.40 – 0.68 | 0.56 | 0.65 |
| CNN_LSTM_F_5 | 0.76 | 0.43 – 0.69 | 0.57 | 0.66 |
| CNN_LSTM_F_7 | 0.74 | 0.29 – 0.66 | 0.53 | 0.64 |

Finally, Table 4 shows the global results by aggregating performance of the respective models across all the tested datasets. Specifically, each row in the second, third, and fourth columns of Table 4 is calculated by averaging the corresponding values of columns three, five, and six, in Table 3. Subsequently, columns five, six, and seven in Table 4 respectively represent the averaged inter-database performance differences between *(i)* the individual models in their respective local testing datasets and their averaged external dataset predictions, *(ii)* the individual models in their respective local testing datasets and the prediction of the ensemble model, and *(iii)* the averaged external dataset predictions of the individual models and the corresponding ensemble model prediction.

*Table 4. Global performance comparison by aggregating results across all datasets: local testing sets using individual models (I), external datasets using individual models (II), and external datasets using an ensemble of individual models (III). The highest absolute values on each column are highlighted in bold. Results are reported in terms of kappa index with respect to the corresponding clinical scorings of reference.*

| Model configuration | Individual models - local dataset (I) | Individual models - external datasets (II) | Ensemble - external dataset (III) | I vs II differences | I vs III differences | II vs III differences |
|---|---|---|---|---|---|---|
| CNN_1 | 0.7417 | 0.5333 | 0.6167 | -0.2083 | -0.1250 | 0.0833 |
| CNN_3 | 0.7367 | 0.5117 | 0.6133 | -0.2250 | -0.1233 | 0.1017 |
| CNN_5 | 0.7217 | 0.4783 | 0.5833 | -0.2433 | -0.1383 | **0.1050** |
| CNN_7 | 0.7133 | 0.4850 | 0.5900 | -0.2283 | -0.1233 | **0.1050** |
| CNN_LSTM_3 | 0.7967 | 0.5433 | 0.6250 | -0.2533 | -0.1717 | 0.0817 |
| CNN_LSTM_5 | **0.8017** | 0.5433 | **0.6267** | -0.2583 | -0.1750 | 0.0833 |
| CNN_LSTM_7 | 0.7733 | 0.5350 | 0.6133 | -0.2383 | -0.1600 | 0.0783 |
| CNN_F_1 | 0.7400 | 0.4683 | 0.5450 | -0.2717 | -0.1950 | 0.0767 |
| CNN_F_3 | 0.7300 | 0.4433 | 0.5350 | -0.2867 | -0.1950 | 0.0917 |
| CNN_F_5 | 0.7167 | 0.4183 | 0.5117 | -0.2983 | -0.2050 | 0.0933 |
| CNN_F_7 | 0.7117 | 0.4150 | 0.5083 | -0.2967 | -0.2033 | 0.0933 |
| CNN_LSTM_F_3 | 0.8000 | 0.4617 | 0.5417 | **-0.3383** | **-0.2583** | 0.0800 |
| CNN_LSTM_F_5 | 0.7983 | 0.4683 | 0.5517 | -0.3300 | -0.2467 | 0.0833 |
| CNN_LSTM_F_7 | 0.7783 | 0.4417 | 0.5333 | -0.3367 | -0.2450 | 0.0917 |

## 4. ANALYSIS OF EXPERIMENTAL DATA data

4.1. Best model (CNN vs CNN-LSTM)

According to Table 4, the proposed deep neural network approach achieves its best generalization performance across all the tested datasets on its *CNN_LSTM_5* architectural variant. This configuration did achieve the best overall performance both in the local as well as in the external dataset prediction scenarios. The implementation of epoch sequence learning by concatenating the LSTM processing block to the output of the preceding CNN feature output layer results on an overall improvement of the model's performance. In general the CNN-LSTM configuration outperforms the respective CNN-only counterpart for the same sequence length at both local and external generalization scenarios. Performance improves with increasing *L*, reaching a saturation value around *L* = 5, after which generalization of the model decreases again below the validation indices obtained for *L* = 3. When using the CNN-only configuration, on the other hand, augmentation of the epoch sequence length does not translate on any network´s prediction improvement. This result for the CNN-only configuration seems to be a consequence of learning overfitting, as Table 1 shows that performance on the respective training sets nevertheless keeps improving with higher values of *L*. For the CNN-LSTM configuration, however, the trend seems to be consistent between the respective training and generalization performances.

4.2. Signal prefiltering

Data from Table 4 seems to rather advise against the use of the optional filtering pre-processing step. A closer look to the results of Tables 1-3, however, does show an inconsistent effect across the individual tested datasets. Actually data could be regarded as inconclusive or even favorable to the use of filters, with the notable exception of the results achieved for the Dublin dataset. As evidenced by data in Tables 2 and 3, the filtering step seems to have a totally different effect on the predictability of this dataset as compared to the rest. Remarkably, however, notice that difficulties of the models in predicting Dublin's data are only evidenced when the validation is carried out on an external prediction scenario. When using Dulin as independent local testing dataset, corresponding data in Table 1 do not show the pronounced performance decay as in the previous setting. This result evidences the database variability problem, and thus importance of expanding the validation procedures beyond the usual local testing scenario, including a sufficiently heterogeneous and independent data sample from a variety of external sources.

4.3. Database generalization performance

Having the expanded validation scenario in mind, and attending to experimental data contained in Tables 1-4, the following general statements might be formulated:

1. The individual model's local-dataset generalization estimation overestimates the actual inter-dataset external generalization. This is a consistent result across all the tested datasets and network configurations (see Table 3). The trend is globally evidenced in Table 4 as well, as *I vs II differences* in the fifth column consistently show negative values. The downgrade in performance when evaluating external data is considerable, with associated kappa indices decreasing on the range between 0.21 up to 0.34 for the tested architectural variants.

2. The proposed ensemble method improves external inter-dataset generalization performance. This result is also consistent across all experimental simulations as evidenced in Tables 3 and 4. The improvement as with respect to the performance of the individual model's estimations ranges between 0.08 and 0.10 on the related kappa indices (see *II vs III differences* in column 7 of Table 4).

3. Individual model's local-dataset generalization estimation still represents an upper bound for the external inter-dataset generalization achieved by the ensemble approach. Similarly evidence is consistent across data of Tables 3 and 4, with absolute kappa differences ranging between 0.12 and 0.26 in this case (*I vs III differences* in column 6 of Table 4).

4.4. Analysis in the context of the expected human performance

Table 5 summarizes literature results reporting on the expected human inter-scorer variability for the sleep staging task. Only works reporting agreement in terms of kappa index are included. Results in Table 5 are structured depending upon if experimentation implements a local or an external validation scenario, enabling a corresponding comparison to our results. In this regard, it is interpreted that a local validation was carried out when agreement among different human scorers belonging to a same center is compared. Usually this also involves the use of their own local database as the source for comparing their scorings. External inter-rater validations, on the other hand, refer to the cases in which a group of experts from different centers compare their scorings using a common independent dataset as reference.

*Table 5. Indices of human inter-rater agreement reported in the literature compared to the performance achieved by the proposed deep-learning approach. Model CNN_LSTM_5 is taken as reference for the values of the automatic approach. Agreement is expressed in terms of kappa index.*

| Dataset | Inter-rater agreement (same center/database) | Our results (local validation) | Inter-rater agreement (different center/database) | Our results (external validation) |
|---|---|---|---|---|
| HMC | 0.74 | 0.79 | --- | 0.63 |
| Dublin | --- | 0.79 | --- | 0.63 |
| SHHS | 0.81-0.83 [47] | 0.82 | --- | 0.61 |
| Telemetry | --- | 0.80 | --- | 0.69 |
| DREAMS | --- | 0.83 | --- | 0.59 |
| ISRUC | 0.87 [45] | 0.78 | --- | 0.61 |
| Other databases | 0.73 [11]<br>0.77-0.80 [48]<br>0.84-0.86 [49]<br>0.86 [50] | --- | 0.46-0.89 [48]<br>0.72-0.75 [51]<br>0.62 [52]<br>0.76 [53]<br>0.68 [54]<br>0.63 [55]<br>0.58 [21]<br>0.75 [50] | --- |
| **Overall (ranges)** | **0.73 – 0.87** | **0.78-0.83** | **0.46 – 0.89** | **0.59-0.69** |

Attending to data in Table 5, our results in the local database generalization scenario are in the range of the expected human agreement under similar conditions (0.78-0.83 ours vs 0.73-0.87 reference). As per dataset the trend holds for HMC (0.79 vs 0.74 reference) and SHHS (0.82 vs 0.81-0.83 reference), while for ISRUC, the automatic system performs somewhat under the expected expert reference (0.78 vs 0.87 reference). For HMC, human reference agreement levels were estimated using a subset of five recordings that were rescored by a total of 12 clinical experts from our sleep lab. The resulting pair-wise kappa agreements between all the combinations of experts were then averaged. To minimize the possibility of a biased case representation, the five recordings were selected, out of the 159 available, using a structured approach based on their relative positioning in the human-computer kappa performance distribution (12.5, 37.5, 50, 62.5 and 87.5 percentiles), where the original clinical expert scorings were used as reference. A similar selection approach was used on a previous study of the authors for the validation of an EEG arousal detection algorithm [56]. No other studies reporting on human kappa agreement were found in the literature for the rest of the datasets used in this work.

As with respect to the external inter-database scenario, analysis of the literature shows a general decrease in human performance when compared to the respective local variability reference. Specifically, two works, [48] and [55], allow comparison between local and external inter-scorer variability on the same dataset. In general results on these works follow the previously mentioned downgrading trend. In [48], however, an exception to this trend is reported in one of the two tested subgroups: 23 recordings scored using the R&K standard, and 21 recordings scored using the AASM rules. Specifically for the first subgroup of 23 recordings, inter-scorer agreement seems to actually increase among scorers coming from different centers ($\kappa = 0.85$-$0.89$ in reference to $\kappa = 0.77$ when scorers belong to the same center). This result seems to represent an outlier, and for the second subgroup the results seem to support again the general downgrading trend reported in the literature ($\kappa = 0.46$-$0.49$ in reference to $\kappa = 0.80$).
Unfortunately human baseline for the external prediction scenario cannot be determined from the current available literature for the databases used in this work. With that in mind, external generalization performance of our automatic scoring approach still seems to fall within the range of the expected human agreement reported for other databases (0.59-0.69 ours vs 0.46-0.89 in general).

4.5. Analysis in the context of other automatic approaches

In Table 6 validation results comprising automatic approaches previously reported in the literature are summarized. As in the previous case, results are structured considering if the performance metrics were obtained on the basis of a local or an external validation scenario. Only studies reporting agreement in terms of kappa index were considered.

*Table 6. Indices of automatic scoring agreement reported in the literature in comparison to the results achieved by the proposed deep-learning approach. Model CNN_LSTM_5 is taken as reference for the performance of our proposed automatic approach. Agreement is expressed in terms of kappa*

| Dataset | Local dataset prediction scenario | Our results (local dataset) | External dataset prediction scenario | Our results (external dataset) |
|---|---|---|---|---|
| HMC | 0.62 [23] | 0.79 | 0.60 [23] | 0.63 |
| Dublin | 0.44 [23]<br>0.84 [18]<br>0.74 [57]<br>0.66 [58] | 0.79 | 0.19 [23] | 0.63 |
| SHHS | 0.65 [23]<br>0.82 [59]<br>0.73 [60]<br>0.83 [61] | 0.82 | 0.62 [23]<br>0.53 - 0.56 [59]<br>0.73 [60] | 0.61 |
| Telemetry | 0.58 [23] | 0.80 | 0.53 [23] | 0.69 |
| DREAMS | 0.62 [23] | 0.83 | 0.43 [23] | 0.59 |
| ISRUC | 0.68 [23] | 0.78 | 0.63 [23] | 0.61 |
| *Overall range (our testing benchmark)* | 0.44 – 0.84 | 0.78 – 0.83 | 0.19 – 0.73 | 0.59 - 0.69 |
| Other databases | 0.86 [62]<br>0.76-0.80 [17]<br>0.84 [63]<br>0.80 [57]<br>0.68 [64]<br>0.81 [60]<br>0.73 – 0.76 [20]<br>0.82 [65]<br>0.77 [58] | --- | 0.42 - 0.63 [48]<br>0.68 – 0.70 [59]<br>0.69 [60]<br>0.72 - 0.77 [21]<br>0.45 - 0.70 [20] | --- |
| *Overall range (all databases)* | 0.44 – 0.86 | 0.78 – 0.83 | 0.19 – 0.77 | 0.59 - 0.69 |

According to Table 6, when comparing local generalization performance on the datasets used in this work, our approach falls within the upper range of the corresponding state-of-the-art results ($\kappa = 0.78 – 0.83$ in this work vs $0.44 – 0.84$ in other works). In particular, the architecture presented in this work clearly outperforms the previous results reported by the authors using the exact same datasets ($\kappa = 0.44 – 0.68$ in [23]). Other works have only reported results in the case of the Dublin and the SHHS datasets. In the first case our approach outperforms existing literature results but in the case of [18] ($\kappa = 0.84$ vs 0.79 in this work). Notice [18] does not report results regarding external independent validation, and therefore overfitting to the local database should not be discarded. In the case of SHHS our approach outperforms the results reported in [60] and matches those in [59]. On another previous author's work [61] slightly better results were reported ($\kappa = 0.83$ vs 0.82 in this work), however as in the previous case, the results in [61] were limited to a local dataset validation scenario. If excluding the previously mentioned studies, no local performance reference has been found in the literature for the rest of the datasets used in this work. Anyhow, when considering the results reported on other benchmarks as a whole, performance still holds on the upper range ($\kappa = 0.44 – 0.86$ globally vs $0.78 – 0.83$ in this work). Notice that the highest performance reported in [62] (κ = 0.86) was obtained using 50% of the data from a small dataset of 8 recordings only, also not including validation data on external datasets.

When considering data on the external dataset validation, Table 6 shows a general global decrease in the performance of the automatic methods as with respect to the corresponding indices on the local database validation scenario. Recall the same trend was reported in the case of human analyses in Table 5. Four works [59] [60] [20] [23] allow comparison between local and external database generalization using the same algorithm. In all of these works performance of the algorithm decreases when tested over external independent datasets. This trend is consistent with the results that we are reporting in our experimentation.

Overall, the highest external database generalization performance reported in the literature has been described in another work [21]. In this work κ between 0.72-0.77 has been reported for the best model on one independent external dataset (IS-RC, see Table 1 in [21]). Local kappa generalization, on the other hand, was not included for the same model among the published results. It is not possible, therefore, to accurately evaluate possible differences between local and external database generalization. Performance comparison in terms of accuracy across different local datasets (see Table 2 in [21]) points out though to significant variability effects, in line with the general trend reported in the literature as mentioned above.

Three studies [59] [23] [60] validate their approach over one of the databases used in this work, namely SHHS. At this respect, our algorithm outperforms the reported external performance in [59] (κ = 0.53-0.56 vs 0.61 in our case). The performance of our method is almost identical to that reported in [23] (κ = 0.62), and it is worse in comparison to that reported in [60] (κ = 0.73). Of these two previous studies only the results in [23] are fully comparable to ours, as they involve the exact same patient selection on SHHS. As stated before, the results in [23] correspond to a previous work of the authors using the exact same datasets as in the present work, but using a different deep neural network architecture. When comparing the average performance of the method presented in [23] across the full set of external datasets used in both studies, it can be shown that the new proposed architecture improves the overall generalization capabilities both in the local (κ = 0.60 in [23] vs 0.80 in this work) as well as in the external (κ = 0.50 in [23] vs 0.63 in this work) validation scenarios.

## 5. DISCUSSION

This study has addressed the extensive validation of a deep-learning based solution for the automatic scoring of sleep stages in polysomnographic recordings. Besides data complexity, a traditional major problem related to the development of automatic sleep staging systems has been the problem of managing all the different sources of variability involved in the decision process. While clinical standard guidelines, such as those contained in the R&K [66] or AASM [1] manuals, aim for a certain level of homogenization, in practice different sources of uncertainty and variability affect the recording and the analysis of the related PSG data (for example, differences in the targeted patient populations, recording methods, or human interpretations, see [23] for a detailed discussion). Validation procedures reported in the literature have been so far limited. Often performance of the method is extrapolated using small or non-independent datasets, mostly composed of data belonging to one particular database only. Consequently, the reported performance is usually bounded to the particular data source, risking overfitting bias. The related validation studies therefore usually lack of the necessary data heterogeneity to allow the establishment of valid generalizations. Our experimentation, together with the analysis of the existing literature, has shown the non-triviality of translating the estimated model's local generalization capabilities to the analysis of independent external datasets. When a system trained with some particular data is presented with similar examples, but gathered from an external database, performance tends to decrease. This result further motivates the necessity of considering external multi-database prediction as a fundamental mandatory step in the validation of this class of systems. It also suggests a critical revision of the related existing literature in this regard.

On this work we wanted to address this issue and confront our design with the broader challenge of evaluating its performance beyond data from a local database partition (local generalization validation). For this purpose we have expanded our tests to include a wide selection of previously unseen external databases (external generalization validation). Intentionally, for this task we have aimed to select databases freely available online (with the only exception of our in-house HMC database) in order to enhance reproducibility of the experiments. Besides our own previous study [23], as far as we know only the recent works of Biswal et al. [60], Bresch et al. [20], and Zhang et al. [59] include public datasets that were evaluated as independent external data, the highest number (three) reported in Zhang et al. [59]. In this study we have included a total of six independent databases, of which five are freely available on the internet, to our knowledge making this study the biggest of its kind.

On this challenging validation scenario, the deep learning architecture proposed in this work has shown good general performance, as compared to both human and automatic references available throughout the literature. We remit to the respective analyses carried out in Sections 4.4 and 4.5. Still, a pertinent remainder is that comparison of the results with other works has to be performed with caution. Effectively, even when referencing the same database source, each study usually involves differences regarding the specific validation approach, the number of involved recordings, or the particular patient conditions in the respective selections. Therefore, only the results provided on an earlier study of the authors [23] can be directly compared, as they address the exact same database benchmark. The specific protocol and subject selection details for the rest of the works analyzed in this study can be found in the referenced publications contained in Tables 5 and 6.

With that in mind, it has been shown that the new CNN+LSTM architecture design introduced in this work translates into considerable improved generalization performance. This improvement has been noticeable on both the local and the external database validation scenarios, and across all the tested configuration variants on the proposed neural network architecture. Experimental data have pointed out as well toward the convenience of adding epoch sequence learning mechanisms using an additional LSTM output block, as with respect to the approach of increasing the length of the input pattern on the CNN-only configuration mode. Moreover, as dimensionality of the CNN input space (4x3000xL) is much bigger than the dimensionality of the LSTM input feature space (50xL), scalability of the solution also improves. Overall, the best performance achieved throughout our experimentation has corresponded to the *CNN_LSTM_5* configuration. No further benefits on increasing the length of the sequence beyond the five epochs have been noticed.

On the other hand, our global results have casted doubt on the convenience of using the proposed signal pre-filtering step. At first sight the result might seem counterintuitive, as filtering was hypothesized to contribute to the input data homogenization, cancelling patient and database-specific artifacts unrelated to the relevant neurophysiological activity, which could hinder generalization of the resulting models. However, data have not shown a consistent effect across all the tested datasets. More research is hence needed to fully understand the underlying causes of the high inter-dataset variability when using the proposed filtering pipeline. The same variability, on the other hand, evidences once again the importance of using a sufficiently heterogeneous and independent data sample, from a variety of external sources, to allow the establishment of valid and generalizable conclusions about the performance of an automatic scoring algorithm.

Last but not least, our experimentation has shown that the use of an ensemble of local models can achieve better generalization performance in comparison with the use of individual local models alone, hence confirming our preliminary results [23]. It is well-known that an effective approach to achieve better generalization of a machine learning model is to increment the amount and heterogeneity of the input training data. The proposed ensemble approach, however, provides advantages in terms of scalability and flexibility of the design [23]. Our results, showing the achievement of better generalization performance, further motivate the exploration of this design principle in future investigations. Notably, the idea of using ensembles resembles the procedure by which expert disagreements are traditionally handled for reaching "consensus". In particular, when assuming that each expert's criterion is equally valid, consensus is usually established resorting to the majority vote [67] [68]. Effectively, each individual local model can be reinterpreted as a "local expert" encapsulating the particular characteristics and ad-hoc knowledge of the human experts on the corresponding source dataset. Statistical, computational, and representational motivations can also be enumerated supporting the use of classifier ensembles [69] [70].

Some possible limitations of our study should be mentioned as well. Specifically, although the proposed ensemble strategy suggests a quantitative improvement in the generalization capabilities among independent databases, there is still notable degradation in the generalization performance in reference to the corresponding local testing datasets. The origin of this degradation must be studied in more detail, investigating alternative approaches to reduce these differences. On the other hand, notice that the analysis of the literature regarding human inter-scorer variability has suggested that differences between local and external validation scenarios are likely to affect human experts in a similar manner. As the goal for an automatic scoring algorithm (when the reference gold standard is based on subjective human scorings) is to achieve comparable agreement with respect to the reference human inter-scorer levels, it remains to be investigated how much of this degradation can actually be explained by the same intrinsic effect in human scoring. For this purpose, the reference levels of expected human agreement, and the corresponding local-external validation differences, need to be assessed for each particular database subject to validation. However, in this study, reference levels of human scoring variability were only available for the HMC, SHHS, and ISRUC databases, and they all regarded local validation scenarios only. Further experimentation is therefore needed involving databases for which the reference levels of human agreement are available including the external validation scenario. Future research will also include the exploration of alternative ensemble combination strategies. The Naive-Bayes combiner [71], for example, might be an appealing approach in taking advantage of the different output probability distributions associated with each individual model in the ensemble.

Future experimentation will also address better hyper-parameterization and data pre-processing methods. In particular, variability of the results for the Dublin dataset with respect to the proposed filtering pipeline remain unclear, and need to be studied in more detail. Among others, further research will be conducted toward addressing the effects of the input sampling rate homogenization (in this study 100 Hz) and contribution of the specific signal derivations selected as input to the model.

**Acknowledgments**

This research did not receive any specific grant from funding agencies in the public, commercial, or not-for-profit sectors